\pdfoutput=1

\documentclass[11pt]{article}

\usepackage{acl}

\usepackage{times}
\usepackage{latexsym}

\usepackage[T1]{fontenc}

\usepackage[utf8]{inputenc}

\usepackage{microtype}

\usepackage{booktabs}
\usepackage{graphicx}
\usepackage{url}

\title{KInITVeraAI at SemEval-2023 Task 3: Simple yet Powerful Multilingual Fine-Tuning for Persuasion Techniques Detection}

\author{Timo Hromadka$^1$ \and Timotej Smolen$^1$ \and Tomas Remis$^1$ \and \\
{\bf Branislav Pecher}$^{2,1}$ \and {\bf Ivan Srba}$^1$  \\
$^1$Kempelen Institute of Intelligent Technologies, Bratislava, Slovakia \\
$^2$Brno University of Technology, Brno, Czechia \\
\texttt{\{timo.hromadka,timotej.smolen,tomas.remis\}@intern.kinit.sk} \\
\texttt{\{branislav.pecher,ivan.srba\}@kinit.sk}}

\begin{document}
\maketitle
\begin{abstract}

This paper presents the best-performing solution to the SemEval 2023 Task 3 on the subtask 3 dedicated to persuasion techniques detection. Due to a high multilingual character of the input data and a large number of 23 labels (causing a lack of labelled data for some language-label combinations), we opted for fine-tuning pre-trained transformer-based language models. Conducting multiple experiments, we find the best configuration, which consists of large multilingual model (XLM-RoBERTa large) trained jointly on all input data, with carefully calibrated confidence thresholds for seen and surprise languages separately. Our final system performed the best on 6 out of 9 languages (including two surprise languages) and achieved highly competitive results on the remaining three languages.
\end{abstract}





\section{Introduction}

The subtask 3 of the SemEval 2023 Task 3 aims at identifying persuasion techniques. The task is a multi-label one, where a model is required to identify which of the 23 persuasion techniques (e.g., an appeal to authority) are present in a given paragraph. The paragraphs are obtained from articles in 6 languages (English, French, German, Italian, Polish, and Russian) collected between 2020 and mid 2022, revolving around widely discussed topics such as COVID-19, climate change, abortion, migration etc. Media sources are both mainstream and alternative news and web portals. Furthermore, the model is tested on 3 surprise languages (Greek, Georgian, and Spanish), for which labeled training data were not available. The importance of the task is eminent --- automatically detected persuasion techniques can be utilized as credibility signals to assess content credibility and thus also to improve disinformation detection. The detailed description of the task is available in \cite{semeval2023task3}.

In this paper, we propose a multilingual system, consisting of a single model tackling all languages. Our main strategy is to fine-tune a large pre-trained transformer-based language model. To find the best performing system, we experimented with different language models (and finally opted for XLM-RoBERTa large due to its performance), hyper-parameter tuning as well as confidence threshold calibration by changing the threshold for prediction in the multi-label classification. We also simulated the zero-shot setting on the training data to adjust the confidence threshold and better estimate the performance of our model on the surprise languages. Furthermore, we experiment with additional configurations, such as translating the data to a single language (English) and using it to fine-tune a monolingual model, applying various text pre-processing strategies, or layer freezing. However, these configurations did not lead to improvements.

Although our system is based on a rather simple concept, it still achieved exceptional results. We ranked 1st for 6 languages (Italian, Russian, German, Polish, Greek and Georgian), 2nd for the Spanish, 3rd for the French and 4th for the English language. In the zero-shot setting introduced by unseen languages, our system also performs exceptionally, achieving the best performance on two languages (Greek and Georgian) and second on the remaining unseen language (Spanish).

Based on our experiments and official ranking on the test set, we make the following observations:

\begin{enumerate}
  \item Combination of a high number of predicted classes and multiple languages (including surprise ones) results in a lack of labeled data, which significantly limits the potential of training multiple monolingual models. Furthermore, even though monolingual models trained on all data translated into English language often achieve state-of-the-art or comparable performance on other multilingual tasks, in this case they are outperformed by the single multilingual models trained on all data. 
  \item Since detecting the presence of persuasion techniques is a complex task (even for humans), the larger models perform significantly better. We also recognized the importance of calibrating the confidence thresholds (for seen and unseen languages separately). At the same time, interestingly, many model configurations (pre-processing, layer freezing, etc.) did not improve model performance.
  \item Even though F1 micro score is the decisive metric in the subtask, we can see a significant difference between F1 micro and macro scores in some of the languages. Similar trend is followed throughout the results of other teams as well. This difference indicates, that our system focuses on the majority classes and struggles with classifying some of the more scarce persuasion techniques within the dataset.
\end{enumerate}

Together with the system description, we also release its source code\footnote{\url{https://github.com/kinit-sk/semeval2023-task3-persuasion-techniques}}, as well as the fine-tuned model used for submitting the final results\footnote{\url{https://huggingface.co/kinit/semeval2023-task3-persuasion-techniques}}.

\section{Background}

The train and dev sets provided by the organizers consisted of 26 663 samples in 6 languages. Each sample consist of a paragraph of news article and zero, one or multiple persuasion techniques (out of 23 possible classes) present in such a paragraph (with the exact span identified). By performing exploratory data analysis of the provided dataset, we observed a high data imbalance in both classes (persuasion techniques) as well as languages (some combinations of classes and languages contain no samples at all) - see Appendix \ref{sec:appendix_data_imbalance}.

The research on the computational propaganda/persuasion techniques detection is still in its early stages, despite its potential importance and utilization for credibility evaluation or disinformation detection~\cite{martinoSurveyComputationalPropaganda2020}. Many existing works closely relate to the SemEval tasks in 2020~\cite{dasanmartinoSemEval2020Task112020a} and 2021~\cite{dimitrovSemEval2021TaskDetection2021}, which preceded the current SemEval 2023 Task 3. The approaches evolved from a simple binary classification of propaganda being present in a document~\cite{barron-cedenoProppyOrganizingNews2019}, through a fine-grained detection of 18 propaganda techniques~\cite{dasanmartinoFineGrainedAnalysisPropaganda2019,dasanmartinoPrtaSystemSupport2020} to detection of 23 persuasion techniques introduced in this task. Moreover, while the methods proposed so far are trained solely on monolingual data, the introduced multilingual data allows to research true multilingual approaches.

\section{System Overview}

The main principle used for development of our system is fine-tuning of a large language model using the data provided within the SemEval task. In similar fashion to other fine-tuning approaches, we add a classification layer at the end of the pretrained model, while also including a dropout layer to prevent overfitting. As input, the language model takes the paragraph, potentially truncated if its length is higher than the maximum input size. No other processing of input is performed (i.e., we are working on paragraph level only). As the task is a multi-class multi-label problem, the predicted output label is not determined based on the maximum probability, but instead by specifying a confidence threshold. All classes that have their probability higher than this confidence threshold are predicted as output label.

\begin{figure*}
    \centering
    \includegraphics[width=1\textwidth]{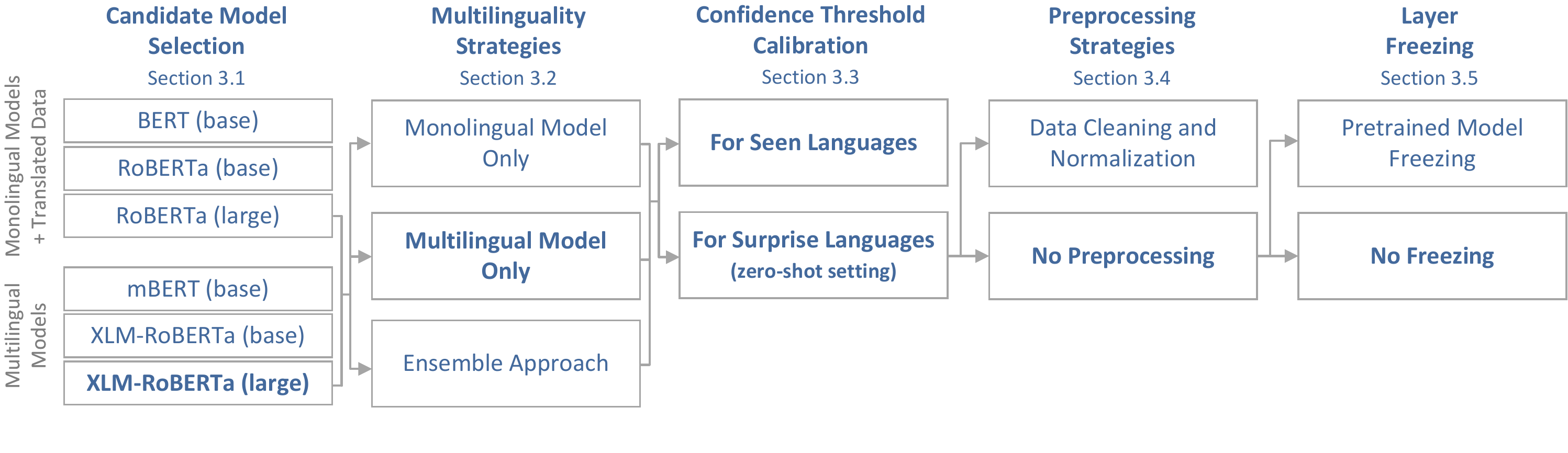}
    \caption{We perform multiple experiments to determine the best configuration for our solution. The performance-improving approaches (denoted in \textbf{bold}) are used in the final solution.}
    \label{fig:system-development-steps}
\end{figure*}

To develop the best configuration of the language model fine-tuning solution, we performed multiple experiments that can be organized into following five steps (which are summarized in Figure \ref{fig:system-development-steps}):

\begin{enumerate}
  \item \textit{Candidate Model Selection}, where we explore the behaviour of both monolingual and multilingual language models on the task, selecting the best performing ones;
  \item  Exploration of \textit{Multilinguality Strategies}, where we compare the best monolingual and multilingual model, and their ensemble;
  \item \textit{Confidence Threshold Calibration}, where we determine the best confidence threshold for both seen languages and surprise languages;
  \item Selection of \textit{Preprocessing Strategies}, where we investigate the benefits data preprocessing;
  \item \textit{Layer Freezing}, where we try different fine-tuning strategies based on what portion of the model is frozen.
\end{enumerate}

The final solution utilizes a single multilingual model fine-tuned on all languages at once, with slightly lowered confidence threshold, with no preprocessing and no layer freezing.

\subsection{Candidate Model Selection}

In the first step, we select the best performing fine-tuned language models separately for monolingual and multilingual models.

At first, the use of monolingual language model has previously shown an exceptional performance even in multilingual setting, where data from non-English languages were translated to English language~\cite{MultilingualPreviouslyFactChecked2023}. Naturally, machine translation, as a specific transfer paradigm for cross-lingual learning~\cite{PIKULIAK2021113765}, may introduce some level of noise and thus break the required correspondence between the original and translated sample (which plays an important role in persuasion techniques detection, since it is especially sensitive to the used wording). At the same time, such approach may potentially better deal with the limited labeled data available for some languages, and even with the zero-shot setting introduced by the surprise languages. Therefore, despite some potential noise introduced by the translation, we decide to explore the behaviour of monolingual models for this task by translating all available data (using Google Translate API) to a single high-resource language (English) and training a monolingual model on such translated data. 
The monolingual models we experiment with are BERT (base) \cite{devlin-etal-2019-bert2} and RoBERTa (base and large) \cite{liu2019roberta}.

On the other hand, the multilingual models provide us with an option to train a single model for all the languages, increasing the amount of available training data. However, multilingual models may lack language-specific understanding required for the complex persuasion technique detection and thus they may not potentially perform as good as the monolingual models. The multilingual models we experiment with are mBERT (base) and XLM-RoBERTa (base and large) \cite{xlmroberta}.

\subsection{Multilinguality Strategies}

In this step, the monolingual, multilingual and ensemble strategy are compared to determine which one is better for persuasion technique detection. Namely, we compare the best performing monolingual model (RoBERTa large), the best performing multilingual model (XLM-RoBERTa large) and their ensemble. The assumption is that the ensemble may exploit the strengths of the combination of translation and monolingual models to deal with the zero-shot setting; and the flexibility of the multilingual model to work with all languages at the same time without a loss of information due to translation. In the ensemble, the predicted output labels from both models are merged together (concatenated) to provide a final prediction.

\subsection{Confidence Threshold Calibration}

In the third step, we perform experiments to determine the best confidence threshold for predicting output classes --- the probability threshold after which the specific class is considered to be the output label for the specific sample. For example, if the threshold is set to 0.2, all classes with predicted probability of at least 0.2 are assigned as predicted labels. To determine the optimal threshold, we use the fine-tuned best-performing monolingual, multilingual model, as well as their ensemble and evaluate their performance on different threshold values. In addition, to determine how our solution will perform on the surprise (unseen) languages, we simulate the zero-shot setting. We randomly select two languages from the training ones as surprise, train all three models on the remaining languages only and use the data from the selected languages only for evaluation. In this way, we are able to better estimate the confidence threshold when working in zero-shot setting on the test set.

Another possibility for the calibration would be to calibrate the confidence threshold for each individual language and class. Although this would improve the performance of the evaluated models on the available data, we believe it would lead to severe overfitting to the distribution of classes on the individual languages and negatively affect the generalizability of our models. Therefore, we opted to pursue the calibration strategy as described in the previous paragraph (single overall calibration for all classes and languages at the same time, with simulated zero-shot setting).

The comparison of all three models for the analysed spectrum of confidence thresholds, also allow us to select the final best-performing model, which is used in the next experiments.

\subsection{Preprocessing Strategies}

In this step, we explore whether preprocessing the data, by removing any potential noise in it, is helpful. To determine the impact of preprocessing, we compare the best-performing model with the already calibrated confidence threshold on the preprocessed data and compare its performance on data without preprocessing. The preprocessing strategies we use are: 1) normalizing white space and punctuation (e.g., reducing multiple punctuation characters to one); and 2) replacing emails, URLs, emojis and hashtags with a placeholder text indicating the specific object (e.g., replacing a specific URL with a generic placeholder ``\{url\}''). We do not evaluate the preprocessing separately, but instead evaluate the model trained on data preprocessed using all strategies at the same time.

\subsection{Layer Freezing}

In the final step, we explore the different layer freezing strategies that can be used during fine-tuning of language models. We compare the best-performing model on following two strategies: 1) no freezing - default setting, where all the layers are fine-tuned (represents the highest level of specialization in the model, as also the layers responsible for generating representations are fine-tuned, but may be more sensitive to overfitting); 2) pretrained layers freezing - in this setting, we freeze 80\% of the pretrained layers and only fine-tune the rest, along with the classification layer (represents a lower level of specialization, mainly in the feature representation).

\section{Experimental Setup}

The only data we use for our system is the official dataset provided for the task \cite{semeval2023task3}. We also use the default training-development split provided for the task. During the development of our solution, we use the development set only for evaluating the different steps and experimental configurations. For the final submission, the language model is fine-tuned on both sets of data. For evaluation purposes we use the F1 micro score, which is the default for this subtask, even though it emphasizes the majority classes over the minority ones.

The different pretrained language models used in our system are chosen from the ones available at Hugging Face. We use the PyTorch deep learning library, version 1.13.1. We have also created a custom pipeline for efficient combining paragraphs and their labels from all articles into a single object, for running all the preprocessing and data translation, and the training of the models.

For each language model, we add a dropout layer with dropout rate of $0.3$, followed by a classification layer with output size of $23$ (one per persuasion technique). Before all the experiments, we perform a hyperparameter optimisation for each language model, mainly focusing on the number of epochs, batch size and learning rate. As starting point for the hyperparameters, we use the values that were determined to perform well in related work (e.g., \cite{zhang_revisiting_2021, mosbach_stability_2021}) and then searched close to these values. The best hyperparameters used for all the language models are: batch size of $16$, ADAM optimizer with $1e-05$ learning rate and fine-tuning for $5$ epochs with early stopping using cross-entropy loss.

\section{Results}

\begin{table*}[tb]
\centering
\small
\caption{Results of different experimental configurations, grouped by the experiment steps (as illustrated in Figure \ref{fig:system-development-steps}). The comparison between monolingual models that utilize translation is provided in the first step (denoted as \textit{Monolingual Model Selection}). The comparison between multilingual models is presented in the second step (denoted as \textit{Multilingual Model Selection}). The best performing models from the first two steps are ensembled and the comparison of this ensemble with the original models is presented in the step 3 (denoted as \textit{Multilinguality Strategies}). The performance of these 3 models on their individual best confidence threshold (which is same for all models) is presented in the step 4 (denoted as \textit{Confidence Threshold Calibration}). For the best performing model from the step 4 (XLM-RoBERTa large with threshold set as 0.29), we report the impact of applying preprocessing strategies (\textit{Preprocessing}) and freezing of 80\% of the pretrained layers (\textit{Layer Freezing}).}
\label{tab:overall-results}
\begin{tabular}{@{}p{3.8cm} l r r @{}}
\toprule
Experiment step         & Configuration                                 & F1 micro (\%)                & F1 macro (\%)              \\ \midrule
Monolingual Model       & BERT (base)                                   & 20.21                        &  7.24                      \\
Selection               & RoBERTa (base)                                & 26.77                        &  5.98                      \\ 
\textit{Section 3.1}    & RoBERTa (large)                               & \textbf{40.38}               & \textbf{15.86}             \\ \midrule
Multilingual Model      & mBERT (base)                                  & 22.06                        &  5.02                      \\
Selection               & XLM-RoBERTa (base)                            & 34.45                        & 13.64                      \\
\textit{Section 3.1}    & XLM-RoBERTa (large)                           & \textbf{45.09}               & \textbf{22.36}             \\ \midrule
\begin{minipage}{0.85\textwidth}
Multilinguality Strategies \\
\textit{Section 3.2}
\end{minipage}          & Ensemble (RoBERTa large + XLM-RoBERTa large)  & \textbf{47.66}               & \textbf{23.99}             \\ \midrule
Confidence Threshold    & RoBERTa (confidence threshold 0.29)           & 45.77                        & 21.88                      \\ 
Calibration             & XLM-RoBERTa (confidence threshold 0.29)       & \textbf{\underline{48.65}}   & \textbf{\underline{27.46}} \\ 
\textit{Section 3.3}    & Ensemble (confidence threshold 0.29)          & 48.15                        & 27.31                      \\ \midrule
\begin{minipage}{0.85\textwidth}
Preprocessing \\
\textit{Section 3.4}
\end{minipage}          & XLM-RoBERTa, threshold 0.29, 2 preprocessing strategies & 48.31                        & 25.83                      \\ \midrule
\begin{minipage}{0.85\textwidth}
Layer Freezing \\
\textit{Section 3.5}
\end{minipage}          & XLM-RoBERTa, threshold 0.29, 80\% freeze  & 36.44                        & 10.57                      \\ 
\bottomrule
\end{tabular}
\end{table*}

The results from the different configurations of our solution are presented in Table \ref{tab:overall-results}. We can observe that the larger language models achieve significantly better performance on the task, both in monolingual setting with translated data (RoBERTa base achieving $26.77\%$ F1 micro compared with RoBERTa large achieving $40.38\%$) and in multilingual setting (XLM-RoBERTa base achieving $34.45\%$ and XLM-RoBERTa large achieving $45.09\%$). In addition, the multilingual setting outperforms the monolingual with translated data. Both of these results can be explained by the complexity of the task, which needs more complex representation provided by the large architectures. At the same time, the nuances needed to correctly detect the persuasion techniques may be lost in translation. However, we can see that both monolingual and multilingual models have their own strengths and weakness, as their ensemble produces the best results overall.

We can also observe significant impact of the confidence threshold calibration. Although the ensemble of monolingual and multilingual models performs best without the calibration, the multilingual model quickly outperforms the ensemble when the best confidence threshold is used for all models (confidence threshold of $0.29$).

Finally, we observe that impact of preprocessing has negligible impact on the overall performance, achieving similar, although slightly lower F1 score. On the other hand, the different layer freezing strategies have significant negative effects on the model, lowering the performance by $\sim12\%$.

\subsection{Confidence Threshold Calibration}
\begin{figure}
    \centering
    \includegraphics[width=0.48\textwidth]{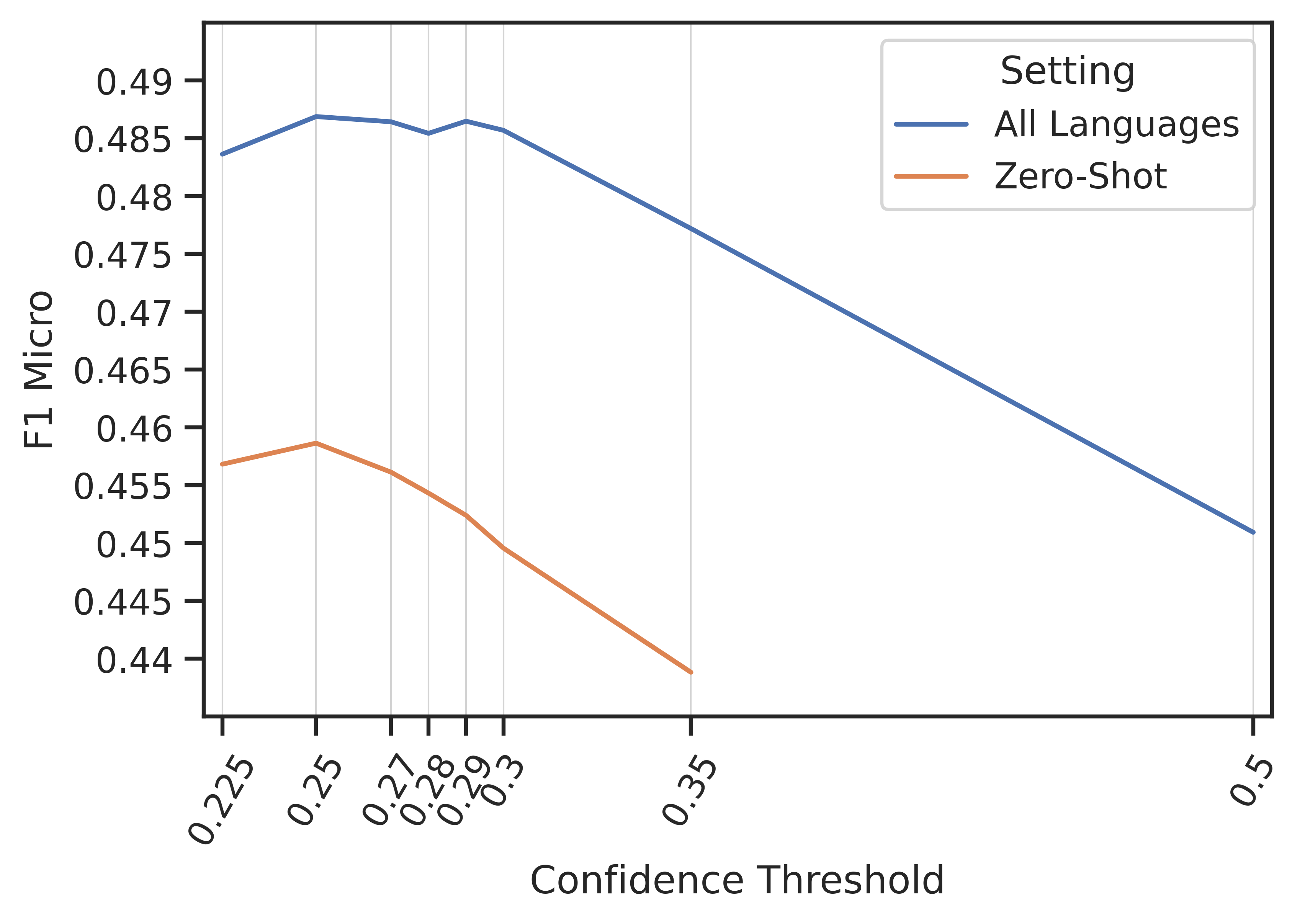}
    \caption{Results (aggregated across languages) from calibrating the confidence threshold for the XLM-RoBERTa (large) model for both the default setting and zero-shot setting.}
    \label{fig:confidence-threshold-calibration}
\end{figure}

The results of the confidence threshold calibration for the XLM-RoBERTa large model in both the default setting (where all languages are seen during training) and zero-shot setting (where some languages are unseen during training) are presented in Figure \ref{fig:confidence-threshold-calibration}. The detailed results for specific languages and for other models are included in Appendix \ref{sec:appendix_threshold_calibration}. 

We observe a significant effect of the calibration, with the performance being significantly higher when lowering the threshold. However, this increase can be observed only to a certain point, after which the performance starts to go down again. The best performing confidence threshold for all models is $0.29$ in the default setting. Changing this threshold also has a significant impact on what model can be considered best. At higher thresholds, the ensemble of monolingual and multilingual models outperforms all others, while at lower values the multilingual model becomes better.

Finally, the best confidence threshold for the zero-shot setting is lower than in the default setting. Instead of the $0.29$, the value of $0.25$ appears to be the best one. This change can be explained by the lower availability of data in this setting, making the models less confident in their predictions. 

\subsection{Final Submission}
The final submission was done using the XLM-RoBERTa large model, trained on the training and development set, using the $0.30$ confidence threshold for seen languages and $0.28$ confidence threshold for the unseen languages (due to lower confidence observed in confidence threshold calibration experiments). Both thresholds are purposefully slightly higher than the best ones found in the experiments, as we expect that training the model on both available data sets will make it also slightly more confident. The official results from the test set are presented in Table \ref{tab:final-results}. Based on the achieved results, our system ranked 1st for 6 languages (Italian, Russian, German, Polish, Greek and Georgian), 2 of which are languages without any training data (Greek and Georgian), 2nd for the Spanish, which is the final unseen language, 3rd for the French and 4th for the English language.

\begin{table}[tb]
\centering
\small
\caption{Results of our system from final submission. $\Delta$ specifies the difference of our results to the best, or the second best (in case we places in the first place) system. The last three languages (Spanish, Greek and Georgian) had no training data. }
\label{tab:final-results}
\begin{tabular}{lccc}
\toprule
Language & F1 micro (\%)    & Rank & $\Delta$   \\ \midrule
English  & 36.157           & 4    & -1.405 \\
Italian  & 55.019           & 1    & +1.140 \\
Russian  & 38.682           & 1    & +0.901 \\
French   & 43.217           & 3    & -3.652 \\
German   & 51.304           & 1    & +0.351 \\
Polish   & 43.037           & 1    & +0.857 \\ \midrule
Spanish  & 38.035           & 2    & -0.071 \\
Greek    & 26.733           & 1    & +0.252 \\
Georgian & 45.714           & 1    & +4.361 \\ \bottomrule
\end{tabular}
\end{table}

\section{Conclusion}

In this paper, we have presented the implementation of the solution proposed by KInITVeraAI team for the subtask 3 within the SemEval 2023 Task 3. Our rather simple, yet powerful, solution utilizes fine-tuning of multilingual language model. In a challenging multi-label task with 23 classes, it achieves very promising performance (F1 micro) of 36-55\% for languages seen during the training, and 26-45\% for unseen languages (zero-shot setting).  

In future, we plan to investigate the potential of prompting and in-context learning on the top of large pre-trained language models (like GTP-3 or ChatGPT). Our hypothesis is that the large size of these models may allow even deeper understanding of the input text. Nevertheless, it will be critical to design appropriate prompts as a part of prompt engineering process, address a potential bias towards majority classes, and also overcome well-known issues with instability of these approaches.

\section*{Acknowledgements}

This research was partially supported by vera.ai project funded by the European Union under the Horizon Europe, GA No. \href{https://doi.org/10.3030/101070093}{101070093}; and by VIGILANT, a project funded by the European Union under the Horizon Europe, GA No. \href{https://doi.org/10.3030/101073921}{101073921}.

\bibliography{anthology,custom}
\bibliographystyle{acl_natbib}

\appendix


\section{Exploratory Analysis: Data Imbalance}
\label{sec:appendix_data_imbalance}

\begin{figure*}[tbh]
    \centering
    \includegraphics[width=1\textwidth]{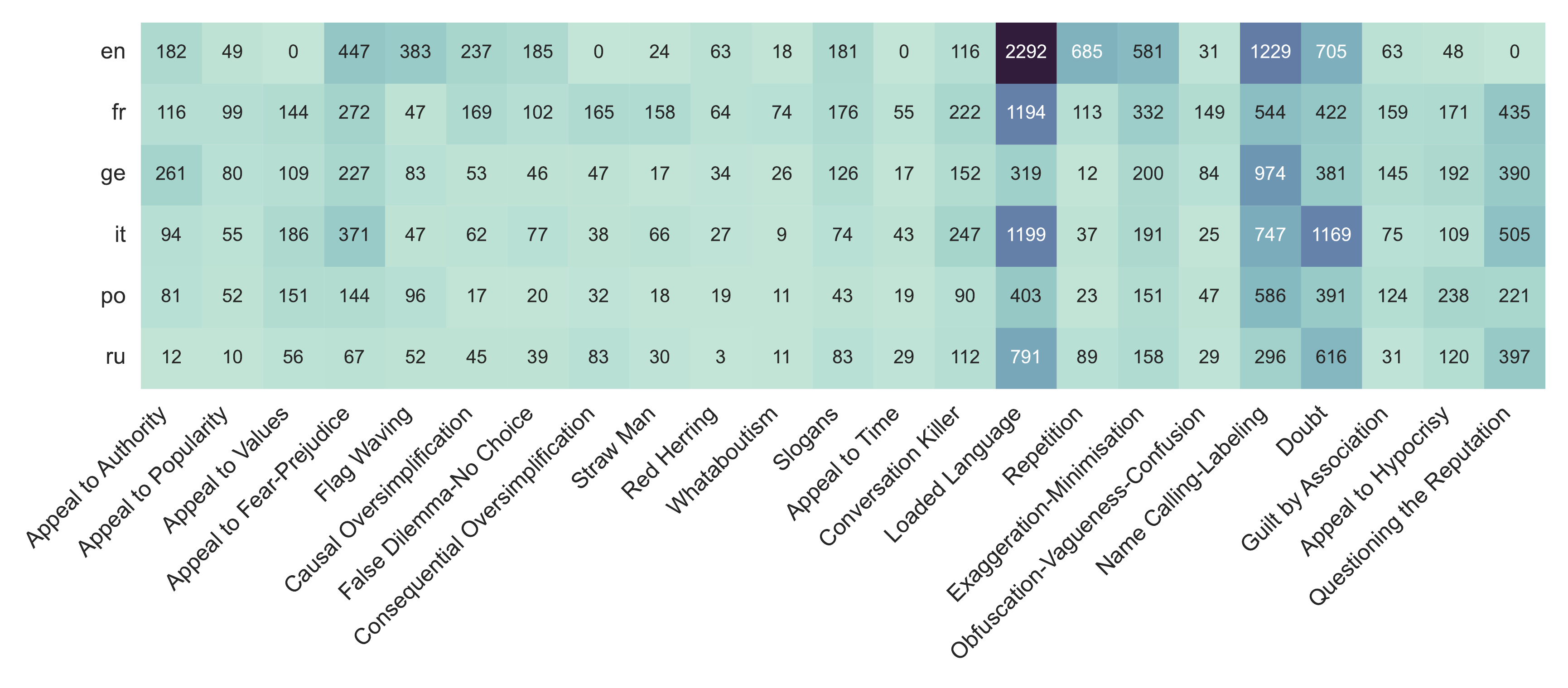}
    \caption{Distribution of labels in the available data sets per language and persuasion technique.}
    \label{fig:eda-class-imbalance}
\end{figure*}

The dataset for persuasion technique detection contains a significant data imbalance, as illustrated in the Figure \ref{fig:eda-class-imbalance}. The data imbalance is present in both the classes, as well as the languages. Although the dataset is working with 23 different persuasion techniques, the technique ``Loaded Language'' is the most frequent one, even representing majority of the labels for some languages. On the other hand, some of the remaining persuasion techniques can even have zero representative samples for some languages, such as ``Appeal to Values'' in English. This imbalance complicates the evaluation of the performance for different models, which is even more augmented by the use of the F1 micro metric that prefers the majority classes.

In addition, the data imbalance is also present in the languages. The number of samples for the English language represent a large portion of the available data. As the task is multilingual, the large representation of the English samples (which also get majority focus in the overall NLP techniques), may have negative impact on the training of a single multilingual model as it may start to prefer English over other languages. Finally, this can also skew the evaluation to prefer models that perform good on the single language, but poorer on the smaller multilingual ones.

\section{Detailed Confidence Threshold Calibration}
\label{sec:appendix_threshold_calibration}

\begin{figure*}[tb]
    \centering
    \includegraphics[trim=2cm 2.5cm 2cm 4.2cm, width=0.9\textwidth]{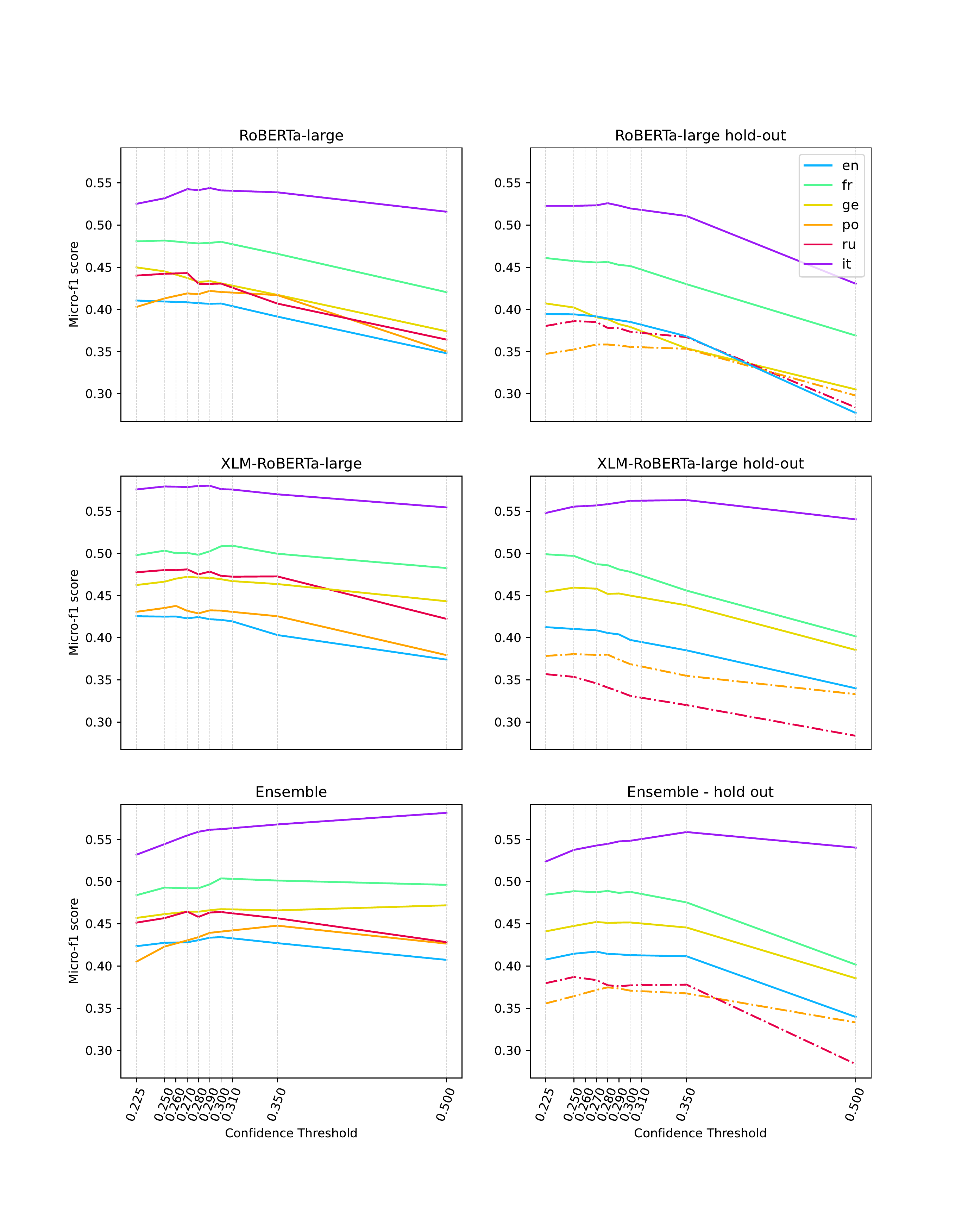}
    \caption{Distribution of labels in the available data sets per language and persuasion technique.}
    \label{fig:confidence-threshold-calibration-detailed}
\end{figure*}

Figure \ref{fig:confidence-threshold-calibration-detailed} depicts a more detailed confidence threshold calibration over different languages and the best performing models identified in this paper (RoBERTa large as monolingual model which uses translation, XLM-RoBERTa model as single multilingual model, and their ensemble). The figure also depicts a comparison between the thresholds when using all languages for training and when working in zero-shot setting with some hold-out languages.

We can observe similar behaviour of the best performing models on the different languages. All models perform the best on the Italian language and the French. For the other languages, we can observe that the RoBERTa models that uses translation performs poorer than the multilingual XLM-RoBERTa. This may be due to the specifics of the other languages, where the translation of the samples obscures some of the details required for the detection of persuasion techniques. 

The same behaviour can also be observed on the threshold. The best threshold determined in aggregate was $0.29$. Looking at individual languages, all of them, except for the French and English, achieve highest performance with this threshold. In case of French, the performance start to drop significantly after confidence threshold value of $0.3$. In case of English, the performance further increases even after the threshold $0.29$ and even achieves the highest performance on the confidence threshold value of $0.225$. This slightly different behaviour on the confidence threshold value may also explain the poorer behaviour of our final model on the French (where we placed 3rd) and English (where we placed 4th).

In addition, we can see a more significant impact of the threshold for the monolingual RoBERTa model than in other models. Reducing the threshold increases the performance more than in other models. However, the performance never overtakes that of the multilingual model or the ensemble of monolingual and multilingual models.

On the zero-shot setting, where we work with lower number of samples for training, the models behave slightly different on the confidence thresholds. For many of the languages seen during training, the best confidence threshold moves more to the left, i.e., lower threshold value provides better performance. We utilize this finding when preparing the final solution. As we train the final system on both training and dev datasets, we have more samples for training and therefore slightly increase the confidence threshold.

We can also observe different impact of the zero-shot setting for different models. The monolingual model that translates the data into English suffer lower decrease of performance than the multilingual model that trains on all the training data in the original languages. However, this behaviour can be expected, as in the monolingual model the unseen languages are still translated and so their samples do not have such importance. However, we still see a significant drop in performance for them. This may point to the fact that the persuasion techniques look slightly different in different languages and this also manifests in translations. On the other hand, the monolingual model suffers more significant overall drop in performance than the multilingual model (where the seen languages still behave with similar performance). This may be due to the lower number of training samples the monolingual model can use, while the drop in number of samples in multilingual model is only in the unseen languages (although it is more significant decrease in number of samples there).

Finally, the behaviour on the unseen languages is also different. We can observe a significant decrease in the performance (as is expected). In addition, the best confidence threshold value is also lower. Instead of the value $0.29$, the best performing one for the unseen languages is $0.25$ on all the models. As the models do not work with any training samples for the specific languages, their confidence is lower, which also lowers the best performing confidence threshold. We also use this finding when preparing the final solution. For the prediction of unseen languages, we use a lower threshold, of value $0.26$ -- the slightly higher value than the best performing one from the experiments is due to the increase in number of training samples, which should also increase the confidence slightly.

\end{document}